\documentclass[lettersize,journal]{IEEEtran}
\usepackage{amsmath,amsfonts}
\usepackage[ruled,linesnumbered]{algorithm2e}
\usepackage{bm} 
\usepackage{varwidth} 
\usepackage{array}
\usepackage{caption}
\usepackage{subcaption}
\usepackage{textcomp}
\usepackage{stfloats}
\usepackage{url}
\usepackage{verbatim}
\usepackage{graphicx}
\usepackage{makecell}

\usepackage{amsthm}

\usepackage{amsmath, amssymb, amsthm}

\usepackage{orcidlink}
\usepackage{amssymb}
\usepackage{multirow}
\usepackage{xcolor}
\usepackage{float}
\usepackage{gensymb}
\usepackage{hyperref}
\usepackage{booktabs}
\usepackage{lettrine}
\usepackage[style=ieee, citestyle=numeric-comp, backend=biber]{biblatex}
\AtBeginBibliography{\small}
\begin{document}

\title{Multi-Head Attention-Based Feature Extractor Integration with Soft Actor-Critic for Porosity Prediction and Process Parameter Optimization in Additive Manufacturing}

\author{
    Kianoush~Aqabakee\,\orcidlink{0009-0008-0450-5360}\thanks{Kianoush Aqabakee is with Department of Electrical Engineering, Amirkabir University of Technology (Tehran Polytechnic), Tehran, Iran, and Department of Mechanical Engineering, Amirkabir University of Technology (Tehran Polytechnic), Tehran, Iran. Email: kianoush.aqabakee@aut.ac.ir.}
    \and
    Leonardo~Stella\,\orcidlink{0000-0002-2670-9873}\thanks{Leonardo Stella is with the School of Computer Science, University of Birmingham, Birmingham B15 2TT, United Kingdom. Email: l.stella@bham.ac.uk.}
}



\maketitle

\begin{abstract}
	Additive manufacturing process optimization requires precise parameter control to minimize defects such as porosity. Traditional reinforcement learning (RL) approaches using discrete action spaces suffer from slow convergence and susceptibility to local optima, limiting their effectiveness for high-precision manufacturing tasks. This study addresses these limitations by employing a continuous action space combined with a novel architecture that integrates a multi-head attention mechanism with the Soft Actor-Critic (SAC) algorithm. The attention-based feature extractor enhances the agent's ability to capture subtle variations in low-dimensional input features, enabling more effective exploration-exploitation balance for navigating value spaces with local minima. We validate our approach on porosity prediction and process parameter optimization in laser powder bed fusion, demonstrating faster convergence and higher final reward values compared to standard RL methods including DQN, PPO, TD3, and vanilla SAC. The proposed methodology achieves a convergence value of 322.79 within 14 episodes, outperforming existing approaches while maintaining stability throughout training.~\footnote{\href{https://github.com/KianoushAqabakee/RL-based-Additive-Manufacturing-Optimization}{\textcolor{red}{\textbf{GitHub Repository}}}}

\end{abstract}


\section{Introduction}\label{sec1}
In recent years, metal additive manufacturing (AM) has gained significant traction across various industries~\cite{10629899}, such as aerospace~\cite{uralde2024novel}, architecture and construction~\cite{zhang2022aerial,9682363}, and biomedical applications~\cite{vyatskikh2018additive,10163841}. The technology presents numerous advantages, including material savings, lower prototyping costs, and the ability to fabricate complex or highly customized components. Despite these benefits, one of the primary challenges hindering the broader adoption of metal AM is ensuring consistency in production~\cite{9347446}. This issue arises from the intricate physical processes involved, the diversity of AM systems, and the choice of metal powders and alloys used~\cite{10292633}. This has led to a growing interest in exploring advanced optimization methods~\cite{9911139,10710827} to enhance process control and improve reliability in AM systems~\cite{9309632,9899412}. Researchers have increasingly turned to data-driven and machine-learning approaches for optimizing metal AM processes to address these challenges.

Traditional CI methods for optimization tasks, such as regression-based techniques and ensemble learning~\cite{su142215475,saf21661}, are limited by the need for large datasets, which make them time and resource intensive in applications such as AM.
In~\cite{9911139}, a cascade forward neural network (NN)-based data-driven approach is introduced to predict process windows and fundamental track geometries in Laser Wire AM of 316L stainless steel. 
Furthermore, in~\cite{8972395}, a convolutional neural network (CNN) is utilized to predict laser power values from melt-pool images, underscoring the broader application of data-driven methods in AM image processing.
In~\cite{10710827}, an automated parameter selection method is presented, leveraging process signatures derived from a Physics-Informed Neural Network (PINN) simulation to optimize Laser Powder Bed Fusion (L-PBF) while obviating the need for repeated Finite Element Method (FEM) simulations.
Similarly, \cite{10549345} introduces an ANN-SOA approach for optimizing metal 3D printing in L-PBF by examining the impact of milling speed, layer thickness, and orientation on surface roughness.

Additionally, heuristic optimization methods, while useful, tend to converge slowly and are prone to getting stuck in local optima, limiting their effectiveness in complex, high-dimensional problems. These limitations underscore the need for more dynamic approaches. RL has emerged as a powerful alternative, offering the ability to optimize complex processes, such as autonomous driving~\cite{10412516}, robotics~\cite{AQABAKEE2025106791}, battery charging management~\cite{HEYDARIANARDAKANI2024111755,10863295}, and AM~\cite{FAIZANMOHAMED2025116377}, by learning directly from interactions with the environment, without the need for extensive pre-existing datasets.

\begin{figure}[H]
	\centering
	\includegraphics[width=.8\linewidth]{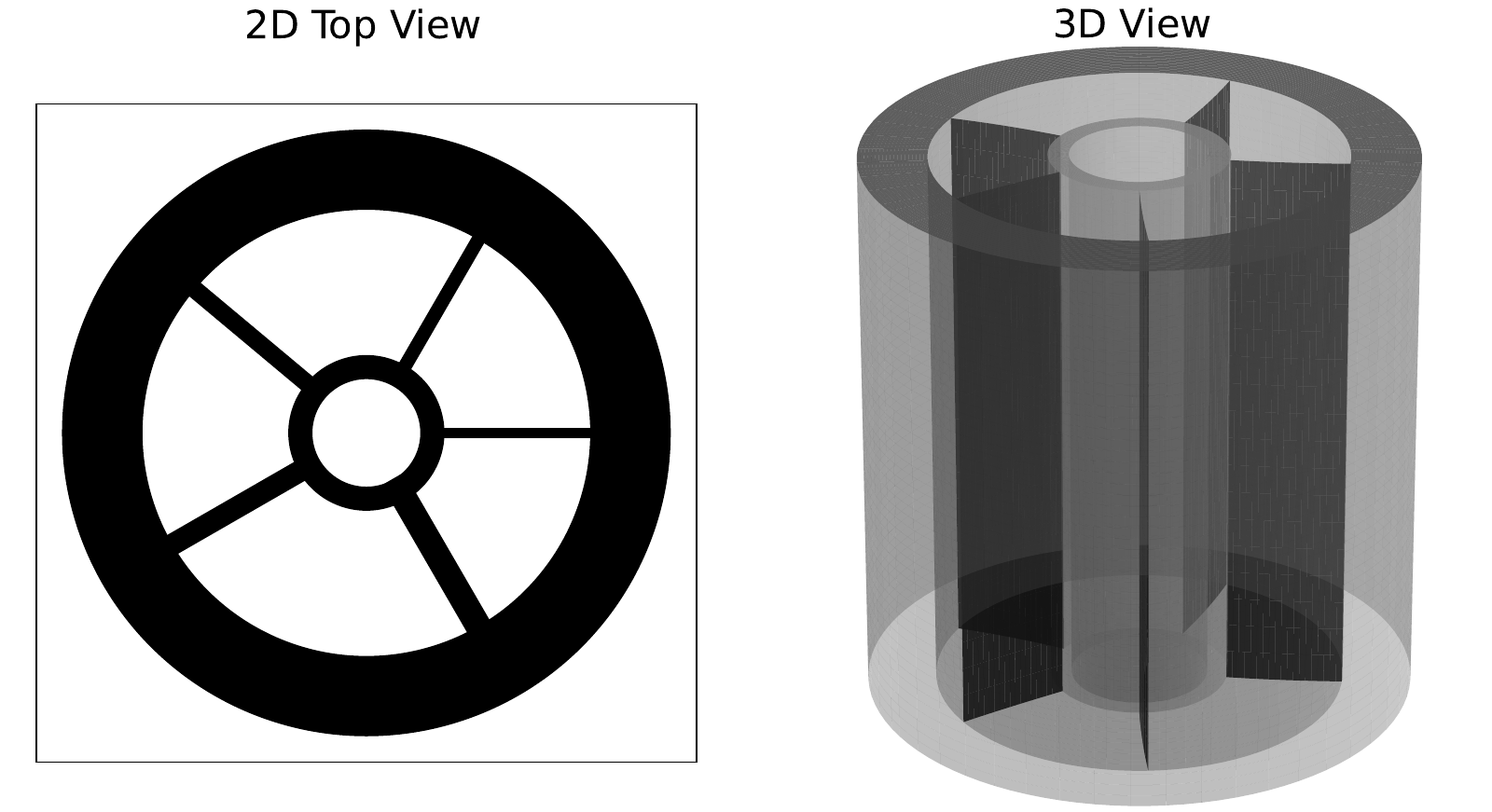}
	\caption{Characterization in production, with different partition thickness for a 3D-printed product.}
	\label{fig:rip}
\end{figure}

In recent years, RL has demonstrated significant potential in optimizing complex systems, including materials science applications. For instance, RL has been applied to process parameter estimation, enabling the adaptation of manufacturing strategies in various component geometries without the need for repeated data collection and model retraining, thereby improving efficiency and cost-effectiveness~\cite{DHARMADHIKARI2023103556}. Additionally, RL has been integrated into automated deposition path planning, where it enhances adaptability in wire arc AM by dynamically adjusting welding parameters to handle complex geometries~\cite{PETRIK202375}. 
Another key application of RL in this domain is the optimization of the processing paths to achieve the desired microstructural properties in materials. In particular, \cite{FAIZANMOHAMED2025116377} explores how RL can be used to control processing conditions and optimize the resulting mechanical properties of materials. 
This study uses a discrete action space due to the RL agent implemented in this research. In contrast, using a continuous action space allows us to include more features by selecting slight parameter variations. These subtle characteristics offered by the continuous action space enable us to enhance our production process. For instance, in~\cite{chrac}, a printed product can have varying partition thicknesses. Figure~\ref{fig:rip} presents an adapted example from this research, illustrating how the approach can be applied to customize the printed product. This strategy helps to customize the printed product. 
Building on this foundation, this paper aims to analyze the most widely used RL approaches and propose an improved algorithm to further advance RL-driven optimization in materials processing.
The contributions of this work are as follows:
\begin{enumerate}
	\item A comparative analysis of various RL approaches is conducted from multiple perspectives to highlight their potential advantages for this specific application.
	\item The integration of a multi-head attention mechanism with a best candidate selection strategy is proposed to achieve an optimal balance between exploration and exploitation, tailored to the value space of this research.
\end{enumerate}

The remainder of this paper is structured as follows: Section~\ref{sec2} details the optimization formulation and the components of the reward function. Section~\ref{sec3} analyzes the relevant value space, followed by the presentation of the proposed methodology. Section~\ref{sec4} reports the implementation results and evaluates their performance. Finally, Section~\ref{sec5} concludes the paper and outlines potential directions for future research.

\section{Preliminaries}\label{sec2}

The primary objective in this RL case is to identify the processing pathways that produce materials with improved mechanical properties performance. The environment is characterized by the state of the material microstructure, and the agent actions represent decisions related to processing parameters. The reward function used in this research is adapted from~\cite{FAIZANMOHAMED2025116377}, which is explained in detail below:
\subsection{Reward Function}
The reward function used in the RL framework was designed by~\cite{FAIZANMOHAMED2025116377} to evaluate the efficacy of each processing step in terms of its contribution to the target mechanical properties of the material. 

The reward function, $R(\hat{\pi}_1, \hat{\pi}_2, \text{VED})$, evaluates the effectiveness of the selected actions in terms of the material microstructural and mechanical properties. The function is formulated on the basis of the Eagar-Tsai thermal model as follows:

\begin{equation}
\label{eq:reward}
	\begin{aligned}
		R(\hat{\pi}_1, \hat{\pi}_2, \text{VED}) = \frac{5}{1 + e^{-6\hat{\pi}_1 - 0.8}} + \frac{5}{1 + e^{-12\hat{\pi}_2 - 0.75}} \\ - \frac{2}{1 + e^{17 - 0.1\text{VED}}} - 2 - \frac{2}{1 + e^{5 - 0.1\text{VED}}},
	\end{aligned}
\end{equation}
where:
\begin{itemize}
	\item $\hat{\pi}_1$: A parameter related to the first aspect of the microstructure or process condition, such as a phase fraction or processing temperature. It can be calculated as follows:
	\begin{equation}
		\pi_1 = \frac{\pi d w}{2 h p},
	\end{equation}
	where d, w, h, p are melt pool depth,  melt pool width, hatch spacing, and powder thickness.
	\item $\hat{\pi}_2$: A parameter related to the second aspect of the microstructure or process condition, possibly representing another phase fraction or time-related factor. It can be calculated as follows:
	\begin{equation}
		\pi_2 = \frac{\gamma d}{v_x \mu \sigma},
	\end{equation}
	where $\gamma, v_x, \mu, \sigma$ are surface tension, horizontal backflow velocity, dynamic viscosity, and laser spot deviation.
	\item $\text{VED}$: A variable that could represent a Volumetric Energy Density (VED) or other energy-related metrics influencing the microstructural development~\cite{buhairireview}.
\end{itemize}

The reward function consists of two main components:
\begin{enumerate}
	\item The first two terms, $\frac{5}{1 + e^{-6\hat{\pi}_1 - 0.8}}$ and $\frac{5}{1 + e^{-12\hat{\pi}_2 - 0.75}}$, model the benefits of achieving desired values for the parameters $\hat{\pi}_1$ and $\hat{\pi}_2$, with higher rewards as these parameters approach optimal values.
	\item The last two terms, $-\frac{2}{1 + e^{17 - 0.1\text{VED}}}$ and $-\frac{2}{1 + e^{5 - 0.1\text{VED}}}$, impose penalties when the energy density, $\text{VED}$, deviates from desirable values. These terms ensure that the energy input remains within an optimal range, thereby discouraging excessive or insufficient energy use.
\end{enumerate}

This reward function encourages the RL agent to optimize the microstructural parameters while considering energy consumption, guiding the learning process toward the desired material properties and efficient processing conditions. 

In the next phase, common RL methods will be analyzed and classified based on their suitability for the problem characteristics, highlighting their advantages and disadvantages.

\section{Methodology}\label{sec3}

\subsection{Value Space}

\begin{figure*}[b]
	\centering
	\begin{subfigure}[b]{0.32\textwidth} 
		\centering
		\includegraphics[width=\textwidth]{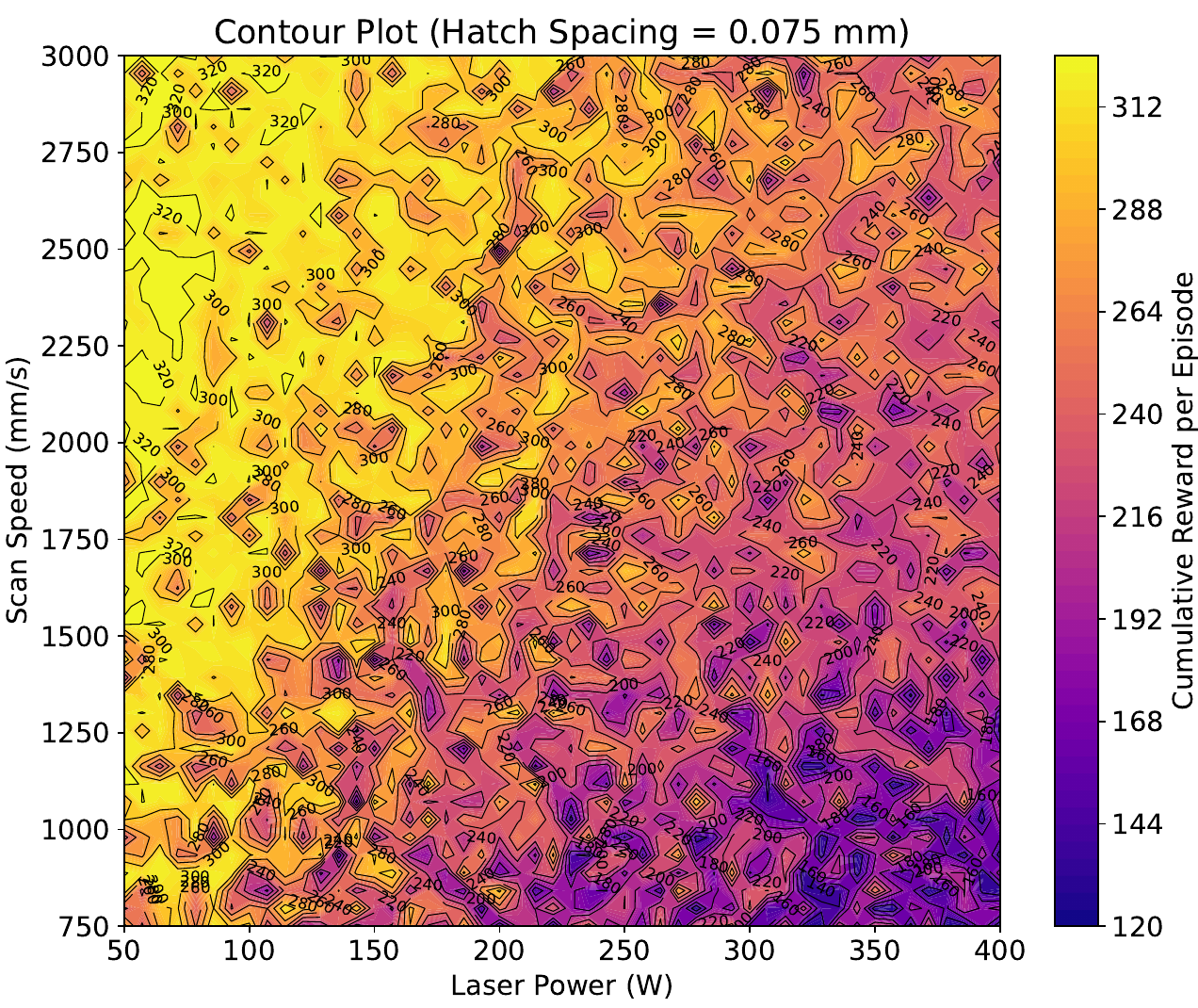}
		\caption{Cumulative Reward per Episode}
		\label{fig:Return_1}
	\end{subfigure}
	\hfill
	\begin{subfigure}[b]{0.32\textwidth} 
		\centering
		\includegraphics[width=\textwidth]{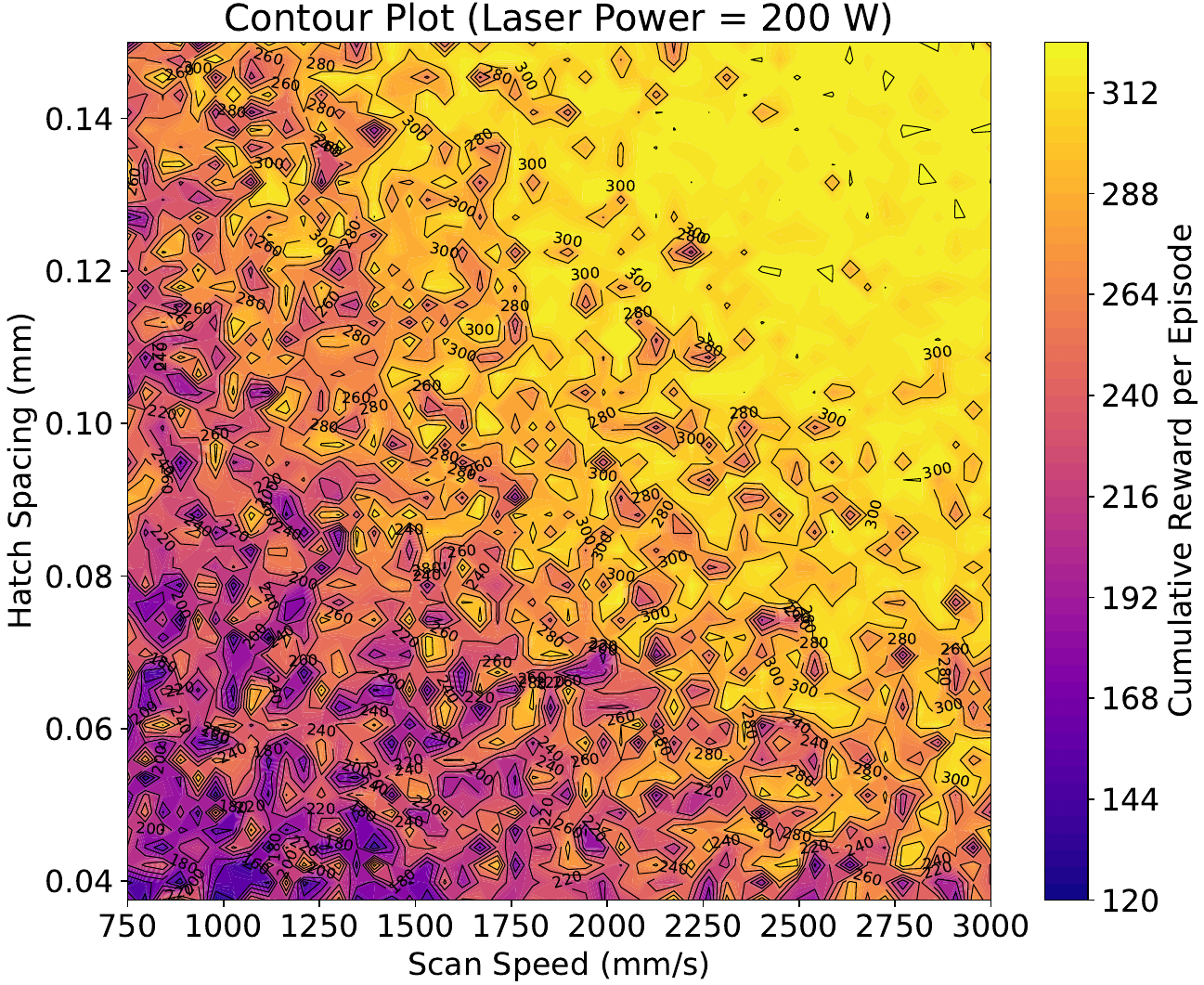}
		\caption{Cumulative Reward per Episode}
		\label{fig:Return_2}
	\end{subfigure}
	\hfill
	\begin{subfigure}[b]{0.32\textwidth} 
		\centering
		\includegraphics[width=\textwidth]{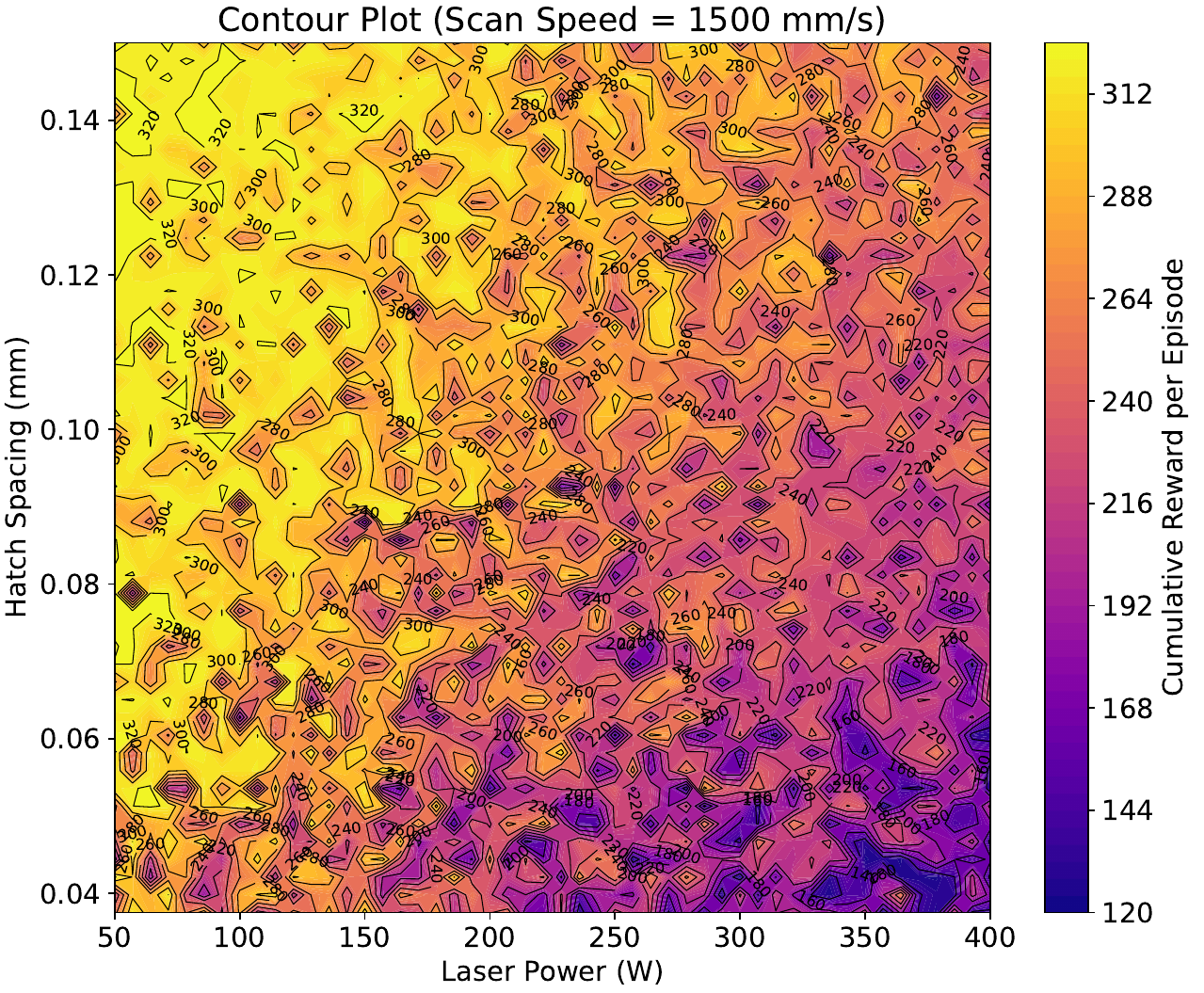}
		\caption{Cumulative Reward per Episode}
		\label{fig:Return_3}
	\end{subfigure}
	\caption{Value Space related to reward function~\ref{eq:reward}}
	\label{fig:three_images}
\end{figure*}

One of the critical factors in identifying the most suitable method for this task is the shape of the value space. Since the task is episodic, the value space can be visualized as depicted in Fig.~\ref{fig:three_images}. 
Three contour plots were generated to analyze the return space based on varying action parameters. In Fig.~\ref{fig:Return_1}, laser power is kept constant, showing the effect of scan speed and hatch spacing on the return. Fig.~\ref{fig:Return_2} fixes the scan speed, illustrating the interaction between laser power and hatch spacing. Lastly, Fig.~\ref{fig:Return_3} presents the return space for a fixed hatch spacing, displaying the combined influence of laser power and scan speed. These plots provide a comprehensive view of how state values shape the return landscape.

The provided figures show that the value space appears to have a smooth topology with small local minima. Therefore, finding the global optimum may not require a high exploration rate and that a lower level of exploration could be sufficient.

In the next step, various DRL approaches are examined, offering different perspectives on the advantages and disadvantages of each with respect to specific characteristics.

\subsection{Characteristics of Deep Reinforcement Learning Algorithms}
In this section, we analyze the characteristics of four selected deep reinforcement learning (DRL) algorithms: Deep Q-Network (DQN)~\cite{dqnref}, Twin Delayed Deep Deterministic Policy Gradient (TD3)~\cite{td3ref}, Proximal Policy Optimization (PPO)~\cite{pporef}, and Soft Actor-Critic (SAC)~\cite{sacref}. 

A discrete action space in RL reduces the optimization complexity, but may overlook nuances in the solution space. Therefore, we also explore continuous RL methods to determine which approach offers better performance. 

By examining their exploration-exploitation strategies, policy types, and convergence behaviors, we aim to identify the most suitable approach for the given task.

\subsubsection{Exploration and Exploitation}
The selected algorithms incorporate different strategies to balance exploration and exploitation. DQN utilizes an $\epsilon$-greedy approach that decays over time, gradually shifting from exploration to exploitation. In contrast, PPO uses a clip objective to limit policy changes, striking a balance between these two aspects. SAC optimizes for maximum entropy, inherently promoting diverse exploration while focusing on high rewards, thus maintaining a dynamic balance.

\subsubsection{Stochastic and Deterministic Policies}
The algorithms can also be distinguished by their policies. DQN and TD3 use deterministic policies; however, DQN directly selects actions based on Q-value approximation without a policy model, whereas TD3 employs a deterministic policy within an actor-critic framework. On the other hand, both PPO and SAC operate with stochastic policies, sampling actions from a probability distribution. SAC further enhances this with entropy maximization, which encourages more extensive exploration.

\subsubsection{Policy-Based and Value-Based Approaches}
DQN is a value-based method among the four algorithms since it approximates the Q-value to determine actions. TD3, PPO, and SAC, however, are policy-based approaches. TD3 and PPO use an actor-critic architecture, where the critic guides policy updates. SAC, also a policy-based method, differs by simultaneously maximizing both reward and policy entropy, thus optimizing for long-term exploration and exploitation.

\subsubsection{Convergence and Robustness}
DQN and TD3 are known for rapid convergence, with TD3 benefiting from deterministic policies, which makes it less robust when dealing with local maxima. PPO and SAC provide more robustness, with PPO’s clipped objective maintaining stable updates, while SAC's entropy maximization allows it to explore diverse action spaces effectively, enhancing its robustness across various environments.
\begin{table*}[t]
\label{tab:rl_comparison}
    \centering
    \renewcommand{\arraystretch}{1.3}
    \caption{Comparison of RL Algorithms}
    \begin{tabular}{lcccc}
        \toprule
        \textbf{Aspect} & \textbf{DQN} & \textbf{TD3} & \textbf{PPO} & \textbf{SAC} \\
        \midrule
        \textbf{Exploration-Exploitation} & $\epsilon$-greedy decay & Target policy smoothing & Clipped objective & Maximum entropy \\
        \textbf{Policy Type} & Deterministic (Q-learning) & Deterministic (actor-critic) & Stochastic & Stochastic (entropy maximization) \\
        \textbf{Method Type} & Value-based & Policy-based & Policy-based & Policy-based \\
        \textbf{Architecture} & Q-network & Actor-Critic & Actor-Critic & Actor-Critic \\
        \textbf{Action Space} & Discrete & Continuous & Continuous & Continuous \\
        \textbf{Convergence} & Fast but unstable & Fast, sensitive to local maxima & Stable updates & Robust, adaptive exploration \\
        \bottomrule
    \end{tabular}
    \label{tab:rl_comparison}
\end{table*}

The discussed exploration-exploitation strategies, policy types, methodological distinctions, and convergence properties are summarized in Table~\ref{tab:rl_comparison}. Based on the analyzed characteristics of the selected RL approaches and the value space, TD3 is expected to achieve the fastest convergence speed and the highest convergence value. However, the presence of small local minima in the value space can be effectively managed with minimal exploration. The SAC method, which offers a more sophisticated and comprehensive exploration strategy, emerges as a strong candidate. Nevertheless, SAC employs a high exploration rate, which can lead to prolonged convergence times. To mitigate this issue, a novel architecture is selected for the actor and critic, allowing the input domain to be mapped into more informative features. This enhancement enables the SAC agent to locate the global optimum more quickly. Consequently, the proposed methodology provides a balanced approach between exploration and exploitation, particularly suited for value spaces with small local minima in their topology.

\subsection{Multi-Head Attention Feature Extraction Mechanism}
\label{MGAFEM}

In this section, we present the architecture of the proposed attention-based feature extractor, which is integrated into the SAC framework. The primary objective of this architecture is to enhance the representation of features by using self-attention, allowing the agent to capture long-range dependencies and better understand the structure of the input space. This is achieved through a sequence of fully connected layers and a multi-head attention~\cite{tensor2tensor, vaswani2017attention} mechanism. Below, we will go over the components of the proposed architecture, starting with the initial fully connected layer, followed by the self-attention mechanism, and concluding with the final fully connected layer.

\begin{figure}[H]
	\centering
	\includegraphics[width=.4\linewidth]{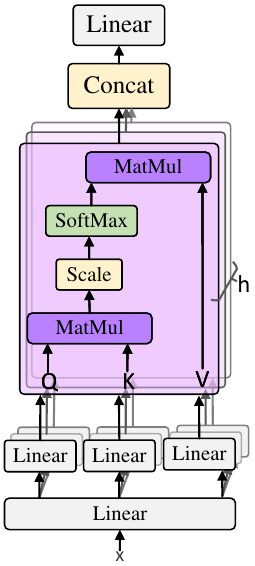}
	\caption{Multi-Head Attention Architecture}
	\label{fig:att_diag}
\end{figure}

\begin{enumerate}
    \item \textbf{Fully Connected Layer 1:}

    Let $ x \in \mathbb{R}^{n} $ be the input observation of dimension $ n $, and let $ W_1 \in \mathbb{R}^{d_{model} \times n} $ be the learned weights. The output of the first fully connected layer is:

    \begin{equation}
    	h_1 = W_1 x + b_1,
    \end{equation}
    where $ b_1 \in \mathbb{R}^{d_{model}} $ is the bias term.
    
    \item \textbf{Self-Attention Mechanism}:

    The self-attention mechanism computes three matrices: query \( Q \), key \( K \), and value \( V \), which are linear projections of the input \( h_1 \).
    \begin{equation}
    	Q = W_Q h_1, \quad K = W_K h_1, \quad V = W_V h_1.
    \end{equation}
    For a single attention head, the attention output is computed as:
    \begin{equation}
    	\text{Attention}(Q, K, V) = \text{softmax}\left(\frac{Q K^\top}{\sqrt{d_k}}\right) V,
    \end{equation}
    where \( d_k \) is the dimensionality of the key.
    
    For multi-head attention with \( h \) heads, the outputs from each head are concatenated and projected using \( W_O \):
    \begin{equation}
        \begin{aligned}
            \text{MultiHead}(Q, K, V) = W_O \left( \bigoplus_{i=1}^{h} \text{Attention}(Q_i, K_i, V_i) \right),
        \end{aligned}
    \end{equation}
    where $\bigoplus$ stands for concatenation.
    \item \textbf{Fully Connected Layer 2}:

    After the attention mechanism, the output is passed through a second fully connected layer, mapping the $d_{model}$-dimensional output to the desired feature dimension \( d \):
    \begin{equation}
    	h_{\text{out}} = W_2 h_2 + b_2,
    \end{equation}
    where \( W_2 \in \mathbb{R}^{d \times d_{model}} \) and \( h_2 \in \mathbb{R}^{d_{model}} \) is the output from the attention mechanism.
\end{enumerate}
The architecture of the proposed model is shown in Fig.~\ref{fig:att_diag}.
%
%
%
%
\begin{figure*}[h]
	\centering
	\includegraphics[width=.8\linewidth]{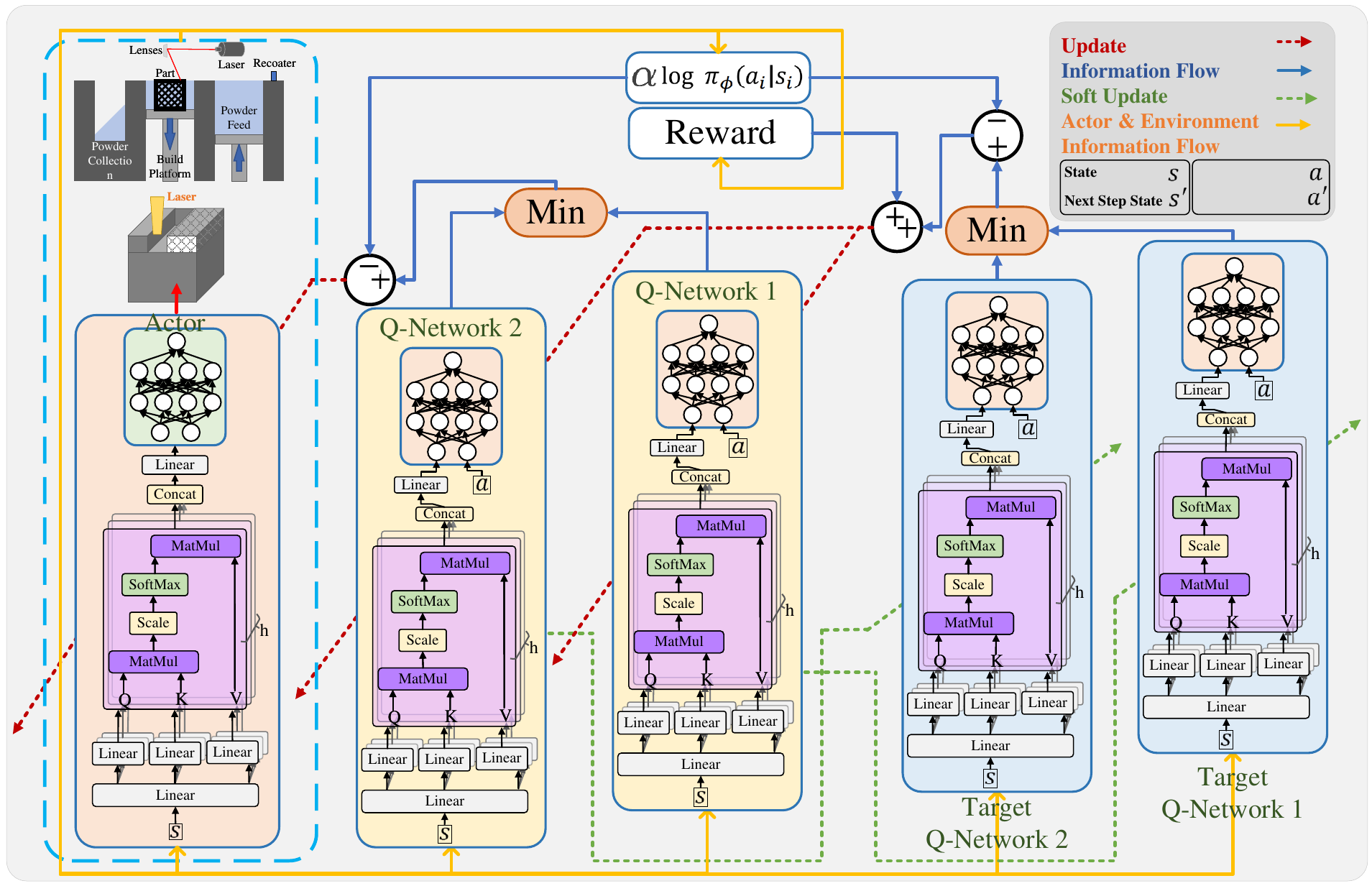}
	\caption{Multi-Head Attention-based Soft Actor-Critic Main Diagram}
	\label{fig:main_diag}
\end{figure*}
\subsection{Multi-Head Attention-based Soft Actor-Critic}
To address the optimization problem based on RL, we define our reward function as~\eqref{eq:reward}. The policy $\pi$ drives the selection of actions, and the corresponding action value function $Q^\pi$, representing the cumulative reward for a given action under $\pi$, is estimated using neural networks. These networks are referred to as the actor and the critic, which are formulated as follows:
\begin{equation}
\label{eq:u_and_Q_explain}
    \left\{\begin{array}{c}
a_t \sim \pi_\theta\left(s_t\right), \\
Q^\pi=Q_\phi\left(\hat{s}_t, \hat{a}_t\right).
\end{array}\right.
\end{equation}
In the SAC algorithm, a dual critic strategy enhances the stability of value estimation by mitigating overestimation bias. Moreover, each critic network has a corresponding target critic network, which is used for more stable target value computation. These networks improve learning efficiency and reduce variance by providing more reliable Q-value estimates during policy updates.
To minimize the error in value estimation, the mean squared error (MSE) is used as the optimization loss function for the critic networks, defined as follows:
\begin{equation}
    \mathcal{L}_{\text {Critic }}=\left(\hat{Q}_{\phi, t}-Q_{\text {target}, t}\right)^2,
\end{equation}
where $ Q_{\text{target}, t}$ denotes the target Q-value computed by the target critic networks. These target networks are updated using a soft update mechanism, providing stable targets for the Q-value updates by decoupling them from the rapidly changing main critic networks, thus mitigating divergence during training. $ Q_{\text{target}, t}$ can be expressed as:
\begin{equation}
    Q_{\text{target}, t} = r_t + \gamma \left( Q_{t+1} - \alpha_t \log \pi(a_{t+1} \mid s_{t+1}) \right),
\end{equation}
where $\alpha$ is the temperature parameter that controls the trade-off between maximizing expected returns and encouraging exploration.
The critic networks provide estimated Q-values that guide the actor network in selecting actions expected to yield higher returns. Accordingly, the actor loss function is defined as:
\begin{equation}
    \label{eq:actor_loss}    
    \begin{aligned}
        \mathcal{L}_{\text{actor}} = - \big( & \min \big( Q_{\phi,1}(s_t, u_t), Q_{\phi,2}(s_t, u_t) \big) \\
        & - \alpha \log \pi(u_t \mid s_t) \big).
    \end{aligned}
\end{equation}
Furthermore, the gradient calculation rule can be obtained as follows:
\begin{equation}
\label{eq:chain_rules}
    \begin{aligned}
        \frac{\partial \mathcal{L}_{\text{actor}}}{\partial \theta} = - \Big[ 
        & \frac{\partial \min \big( Q_{\phi,1}(s_t, u_t), Q_{\phi,2}(s_t, u_t) \big)}{\partial \hat{a}_t} 
        \frac{\partial \hat{a}_t}{\partial \pi_\theta(s_t)} \\
        & - \alpha_t \frac{\partial (\log \pi(a \mid s))}{\partial \pi_\theta(s_t)} 
        \Big] \frac{\partial \pi_\theta(s_t)}{\partial \theta}.
    \end{aligned}
\end{equation}
Additionally, the temperature parameter \(\alpha\) is updated using the following objective:  
\begin{equation}
    \mathcal{L}_{\alpha} = -\alpha \big( \log \pi(a \mid s) + \mathcal{H}_{\text{target}} \big),
\end{equation}
where $\mathcal{H}_{\text{target}}$ is target entropy. This loss ensures that the entropy temperature $\alpha$ is automatically adjusted to balance exploration and exploitation by driving the policy entropy toward the desired target $\mathcal{H}_{\text{target}}$ Finally, the target critic networks are updated using a soft update process, based on the critic parameters. This is formulated as:
\begin{equation}
    \phi_{\text{target}} = \tau \phi_{\text{target}} + (1-\tau) \phi_{\text{target}},
\end{equation}
where \( 0 < \tau < 1 \). The architecture of the actor and all four critic networks is based on the Multi-Head Attention Feature Extraction Mechanism, as described in~\ref{MGAFEM}. The entire mechanism is illustrated in Fig.~\ref{fig:main_diag}. 
\section{Results and Discussion}\label{sec4}

\begin{figure}[h]
	\centering
	\includegraphics[width=1\linewidth]{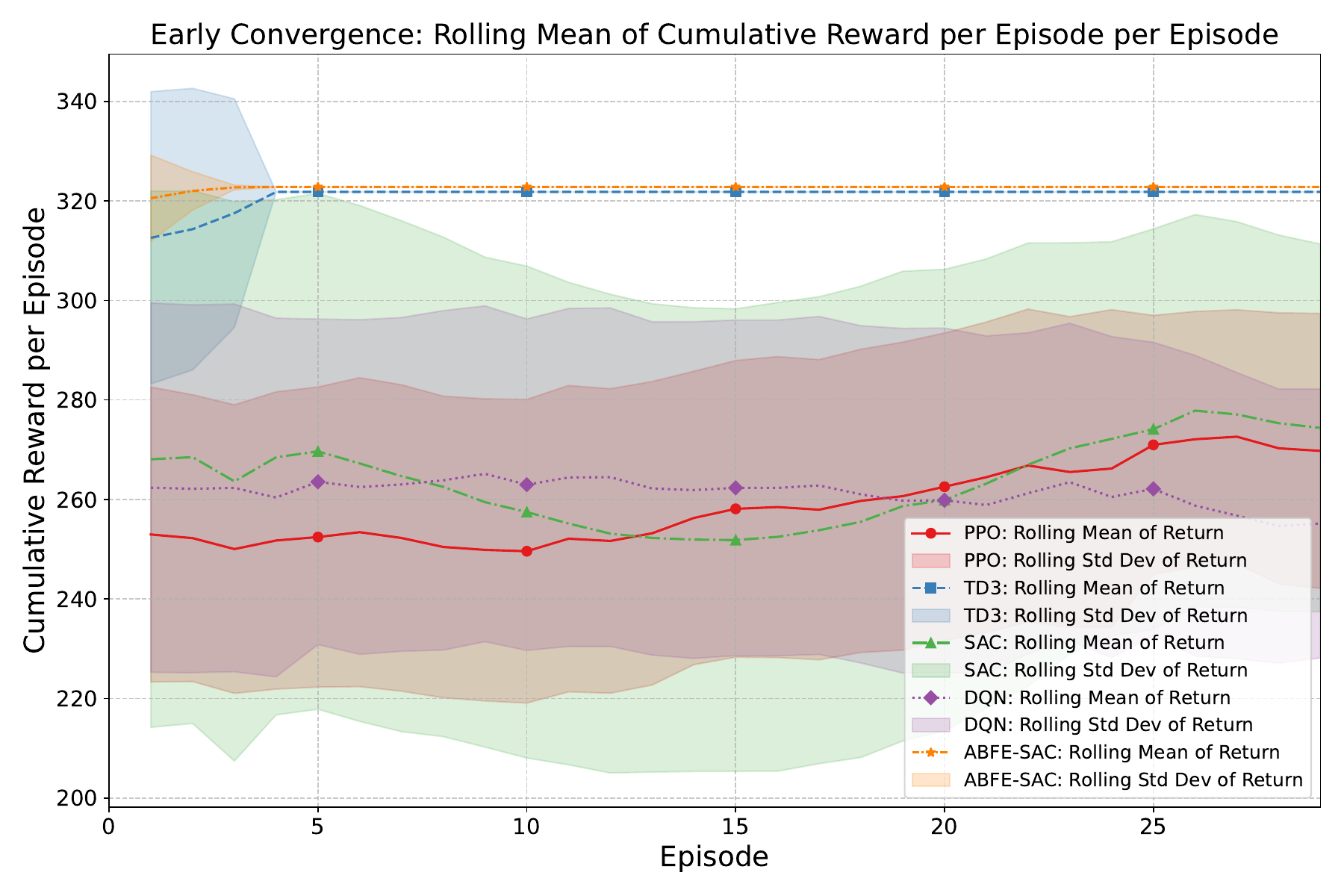}
	\caption{Return diagram rolling window size 5}
	\label{fig:return_plot_early}
\end{figure}

\begin{figure}[h]
	\centering
	\includegraphics[width=1\linewidth]{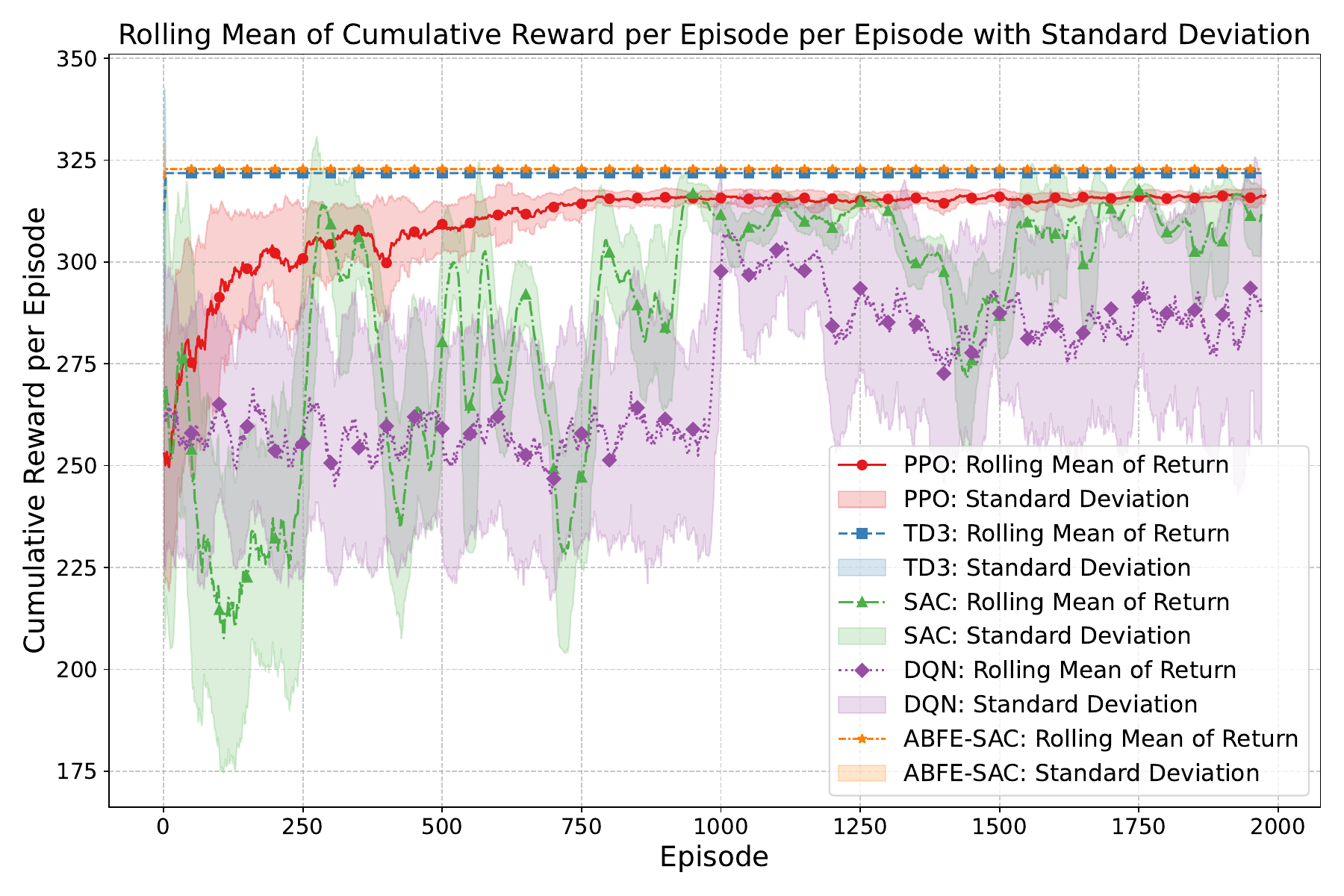}
	\caption{Return diagram rolling window size 30}
	\label{fig:return_plot}
\end{figure}

The discussed algorithms were implemented over a total of 2000 episodes. A discount factor of 0.99 was used across all simulations. For PPO, the clip range was set to 0.2, GAE lambda to 0.95, and the entropy coefficient to zero. In TD3, the soft update rate was 0.005, policy noise was applied, and the policy update delay was set to 2. For SAC, the soft update rate was 0.005, with automatic tuning of the entropy coefficient and a target update interval of 1. The remaining hyperparameters used in this implementation are detailed in Table~\ref{tab:hyperparameters}. 
\begin{table}[h]
	\centering
	\small
	\caption{Hyperparameters used in implementation for the discussed RL methods.}
	\label{tab:hyperparameters}
	\begin{tabular}{@{}lcccc@{}}
		\toprule
		\textbf{Method} & \textbf{Learning Rate} & \textbf{Buffer Size} & \textbf{Batch Size} \\ \midrule
		DQN & 1e-4 & 1e6 & 32 \\
		PPO & 3e-4 & - & 128 \\
		TD3 & 1e-3 & 1e6 & 128 \\
		SAC & 3e-4 & 1e6 & 256 \\
		ABFE-SAC & 3e-4 & 1e6 & 64 \\ \bottomrule
	\end{tabular}
\end{table}
The cumulative reward (Return) for the first 30 episodes of the training process is depicted in Fig.~\ref{fig:return_plot_early}. As observed, the initial values for the DQN and PPO methods fall within a specific range, indicating a balanced trade-off between exploration and exploitation. In contrast, the SAC method demonstrates a higher level of exploration, as evidenced by its substantial deviation and oscillation. 
The TD3 and the proposed method display minimal deviation, with the proposed method achieving slightly better convergence and a faster convergence speed than TD3.
Furthermore, the complete simulation results are presented in Fig.~\ref{fig:return_plot}. Over the course of long-term training, the DQN algorithm demonstrates the weakest convergence, with the deviation in the return diagram remaining consistent from the beginning to the end of training. The SAC algorithm exhibits the highest degree of exploration during the initial episodes. However, after a substantial number of episodes, the deviation decreases, indicating the onset of convergence. This characteristic makes SAC a promising choice in environments where the return space contains large local maxima. The PPO algorithm, by contrast, demonstrates a smooth convergence pattern. It initially displays high deviation due to exploration but steadily converges to a high return value with minimal and bounded deviation. 
Finally, the TD3 algorithm shows stable and high convergence, though the proposed method achieves slightly better final performance and converges a bit faster than TD3.


Table.~\ref{tab:convergence_values_times} presents the convergence values and convergence episodes for discussed methods, illustrating the point at which each method achieves stable performance.
\begin{table}[h]
	\centering
	\small
	\caption{Convergence values and convergence times for different RL methods.}
	\begin{tabular}{lcc}
		\toprule
		\textbf{Method} & \makecell{\textbf{Convergence} \\ \textbf{Value}} & \makecell{\textbf{Convergence} \\ \textbf{Episodes}} \\
		\midrule
		DQN & $288.65$ & $-$ \\
		PPO & $316.26$ & $944$ \\
		SAC & $312.45$ & $1293$ \\
		TD3 & $321.80$ & $14$ \\
		ABFE-SAC & $322.79$ & $14$ \\
		\bottomrule
	\end{tabular}
	\label{tab:convergence_values_times}
\end{table}
Ultimately, the proposed algorithm demonstrates the best performance across all criteria.

\begin{figure}[h]
    \centering
    \begin{subfigure}[b]{1\linewidth}
        \centering
        \includegraphics[width=.8\textwidth]{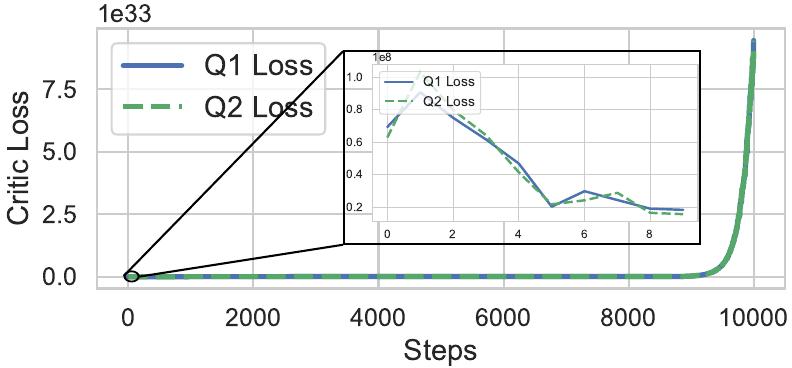}
        \caption{Critic loss.}
        \label{fig:SAC_Res_criticE}
    \end{subfigure}
    \hfill
    \begin{subfigure}[b]{1\linewidth}
        \centering
        \includegraphics[width=.9\textwidth]{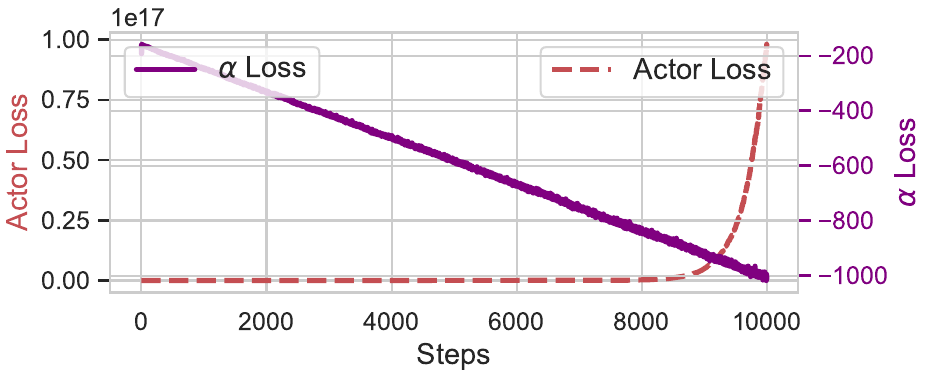}
        \caption{Actor and entropy temperature losses.}
        \label{fig:SAC_Res_ActorAlphaLoss}
    \end{subfigure}
    \hfill
    \begin{subfigure}[b]{1\linewidth}
        \centering
        \includegraphics[width=.8\textwidth]{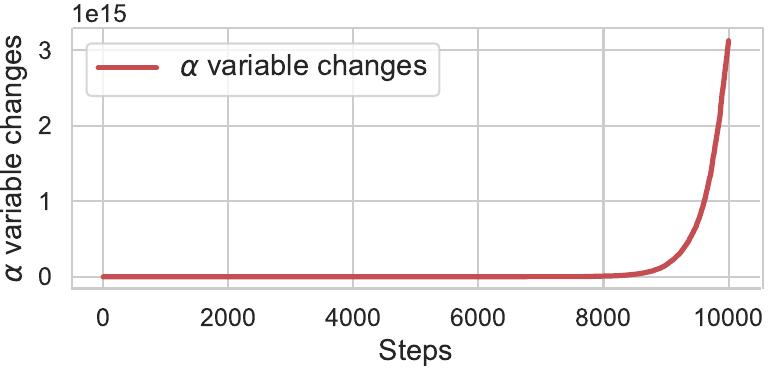}
        \caption{The entropy temperature.}
        \label{fig:Results_alpha}
    \end{subfigure}
    \hfill
    \begin{subfigure}[b]{.8\linewidth}
        \centering
        \includegraphics[width=1\textwidth]{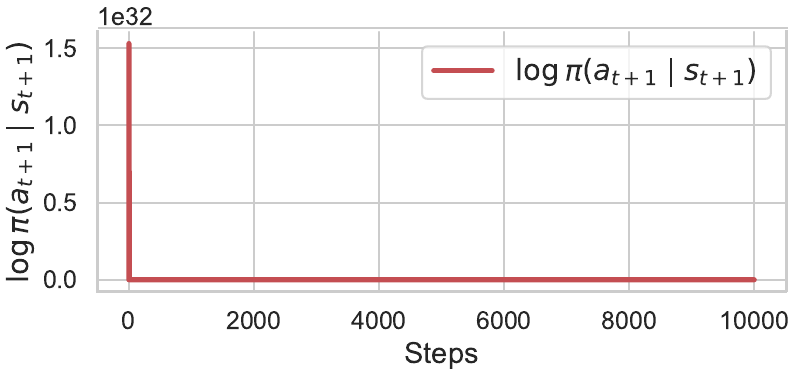}
        \caption{The policy log likelihood.}
        \label{fig:SAC_Res_logpi}
    \end{subfigure}
    \hfill
    \begin{subfigure}[b]{1\linewidth}
        \centering
        \includegraphics[width=.8\textwidth]{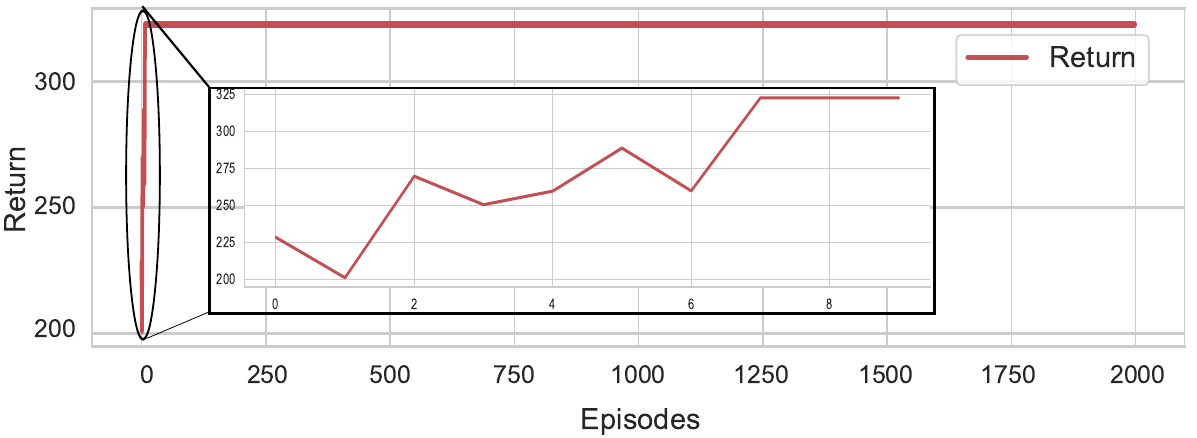}
        \caption{Cumulative Reward per Episode.}
        \label{fig:SAC_Res_reward}
    \end{subfigure}
    \caption{Proposed Multi-Head Attention-based SAC agent results.}
    \label{fig:temp}
    \vspace{-10px}
\end{figure}

The final aspect to be analyzed is whether the agent successfully converged to the global solution. To provide further insight into the training process, Fig.\ref{fig:temp} illustrates various components of the agent. The return value stabilizes at the convergence point after approximately six episodes, as depicted in Fig.\ref{fig:SAC_Res_reward}. Additionally, the value space estimation error, computed by the critic networks, initially exhibits a decreasing trend, as shown in Fig.\ref{fig:SAC_Res_criticE}. Similarly, the policy log-likelihood declines over time, as illustrated in Fig.\ref{fig:SAC_Res_logpi}.

Following the initial convergence phase, the agent attempts to increase the weight of the entropy term to enhance exploration and identify actions with higher probabilities, thereby avoiding local minima. Consequently, the loss associated with the $\alpha$ parameter drives an increase in $\alpha$, as shown in Fig.\ref{fig:Results_alpha}. This results in a subsequent rise in the actor error, as depicted in Fig.\ref{fig:SAC_Res_ActorAlphaLoss}. However, after many training episodes, the return value remains stable, indicating that the agent has successfully reached the global solution.

\section{Conclusion}\label{sec5}

%

This study explores the porosity prediction to optimize the parameters of AM processes. We propose a RL methodology with two key innovations. First, we incorporate both reward and entropy maximization to identify the global optimum in the value space effectively. Second, we enhance the convergence speed and value by using a continuous action space and integrating a multi-head attention feature extractor into the SAC algorithm. 
This allows us to capture the thin features and enhance sensitivity to minor variations. As a result, the need for practical testing is reduced, leading to improved material characterization.

In addition, we evaluate standard RL approaches to provide a comparative analysis, highlighting their respective characteristics and contrasting them with the performance of the proposed method. The comparison criteria focus on the convergence speed and the final convergence value, which shows the superior performance of the proposed approach.

This method can easily incorporate data directly coming from machines and can be used with much richer data sets.
Future research could investigate the extension of the proposed methodology to other manufacturing techniques, such as directed energy deposition, in addition to its current application in laser powder bed fusion. Furthermore, future work may address challenges related to instability and nonstationarity avoidance~\cite{zhang2025experiencereplayinnovativedynamics}, as well as the development of optimization strategies for multi-agent systems~\cite{10383787}.

\printbibliography

\begin{IEEEbiography}[{\includegraphics[width=1in,height=1.25in,clip,keepaspectratio]{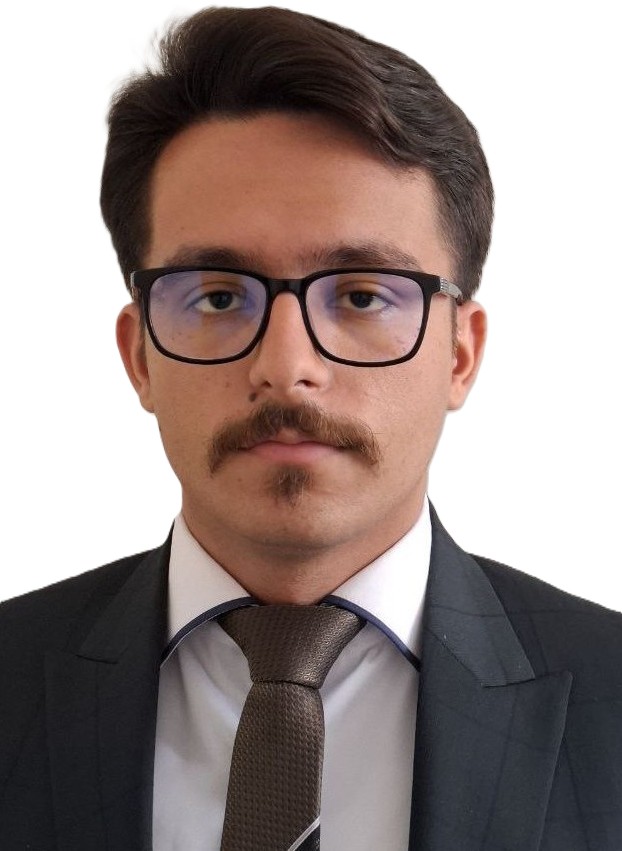}}]{Kianoush Aqabakee}
received the B.S. degree in Electrical Engineering from the Department of Electrical Engineering, Ferdowsi University of Mashhad, Iran, in 2021, and the M.S. degree in Mechatronics Engineering from the Departments of Electrical Engineering and Mechanical Engineering, Amirkabir University of Technology, Tehran, Iran, in 2024. Since 2025, he has been a PhD student in the Department of Information Technology, Faculty of Engineering and Architecture, Ghent University, Belgium.

His research interests include computational intelligence, control theory, and biomedical engineering.
\end{IEEEbiography}

\begin{IEEEbiography}[{\includegraphics[width=1in,height=1.25in,clip,keepaspectratio]{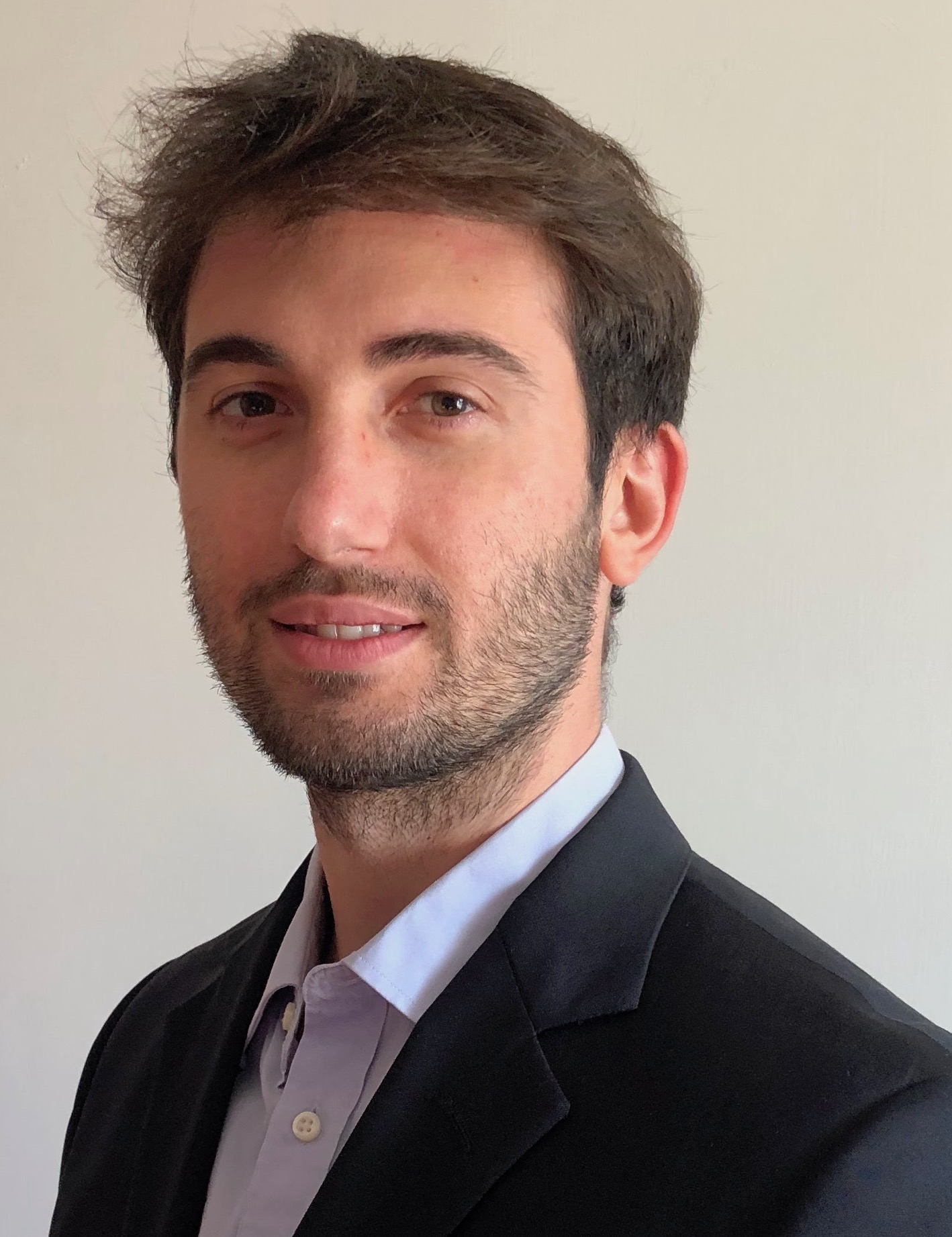}}]
{Leonardo Stella} received the Laurea Triennale degree in Computer Engineering in 2013 from the Universit\`a di Palermo, Italy, after being awarded the departmental scholarship to study for two years at USI, Universit\`a della Svizzera Italiana, Switzerland. He received the Laurea Magistrale degree in Artificial Intelligence and Robotics in 2016 from the university La Sapienza, Italy, and the PhD degree in 2019 from the Department of Automatic Control and Systems Engineering, University of Sheffield, United Kingdom, under the supervision of Prof. Dario Bauso and Dr. Roderich Gro\ss. He was awarded the departmental scholarship for the full duration of the doctorate. Since 2022, he is Assistant Professor in the School of Computer Science, University of Birmingham. From 2019 to 2022, he was Lecturer in the Department of Computing, University of Derby. He has been Guest Editor of International Journal of Robust and Nonlinear Control (IJRNC).

His research interests are in the areas of game theory, control, multi-agent learning, and their applications. Examples include theoretically grounded multi-agent learning, biology, process parameter optimization in materials science.
\end{IEEEbiography}

\end{document}